\documentclass[journal]{IEEEtran}
\IEEEoverridecommandlockouts

\usepackage{lipsum}
\usepackage{booktabs}
\usepackage{pgfplots}
\pgfplotsset{compat=newest}

\usepackage{enumitem}
\usepackage{array}
\usepackage{multirow}
\usepackage{rotating}
\usepackage{caption}
\usepackage{subcaption}
\usepackage{adjustbox}
\usepackage{cite}
\usepackage{hyperref}
\usepackage{amsmath,amssymb,amsfonts}
\usepackage{algorithmic}
\usepackage{graphicx}
\usepackage{textcomp}
\usepackage{xurl}
\usepackage{svg}
\usepackage{xcolor, colortbl}

\def\BibTeX{{\rm B\kern-.05em{\sc i\kern-.025em b}\kern-.08em
    T\kern-.1667em\lower.7ex\hbox{E}\kern-.125emX}}
    
\newcommand{\printfnsymbol}[1]{%
  \textsuperscript{*}%
}    
\definecolor{backgroundcolor}{HTML}{d2d2d2}
\definecolor{questioncolor}{HTML}{3167BD} 
\definecolor{answercolor}{HTML}{000000}
\renewcommand\fbox{\fcolorbox{black}{backgroundcolor}}

\begin{document}

\title{How Good is ChatGPT at Face Biometrics? \\ A First Look into Recognition, \\ Soft Biometrics, and Explainability}

\author{Ivan DeAndres-Tame, Ruben Tolosana, Ruben Vera-Rodriguez, \\Aythami Morales, Julian Fierrez, Javier Ortega-Garcia \\

Biometrics and Data Pattern Analytics - BiDA Lab, Universidad Autonoma de Madrid \\
{\tt\small \{ivan.deandres, ruben.tolosana, ruben.vera, \\aythami.morales, julian.fierrez, javier.ortega\}@uam.es}
}

\maketitle

\begin{abstract}
Large Language Models (LLMs) such as GPT developed by OpenAI, have already shown astonishing results, introducing quick changes in our society. This has been intensified by the release of ChatGPT which allows anyone to interact in a simple conversational way with LLMs, without any experience in the field needed. As a result, ChatGPT has been rapidly applied to many different tasks such as code- and song-writer, education, virtual assistants, etc., showing impressive results for tasks for which it was not trained  (zero-shot learning).

The present study aims to explore the ability of ChatGPT, based on the recent GPT-4 multimodal LLM, for the task of face biometrics. In particular, we analyze the ability of ChatGPT to perform tasks such as face verification, soft-biometrics estimation, and explainability of the results. ChatGPT could be very valuable to further increase the explainability and transparency of automatic decisions in human scenarios. Experiments are carried out in order to evaluate the performance and robustness of ChatGPT, using popular public benchmarks and comparing the results with state-of-the-art methods in the field. The results achieved in this study show the potential of LLMs such as ChatGPT for face biometrics, especially to enhance explainability. For reproducibility reasons, we release all the code in GitHub\footnote{\url{https://github.com/BiDAlab/ChatGPT_FaceBiometrics}}.
\end{abstract}



\begin{figure*}[t]
    \centering
    \includegraphics[width=0.90\linewidth]{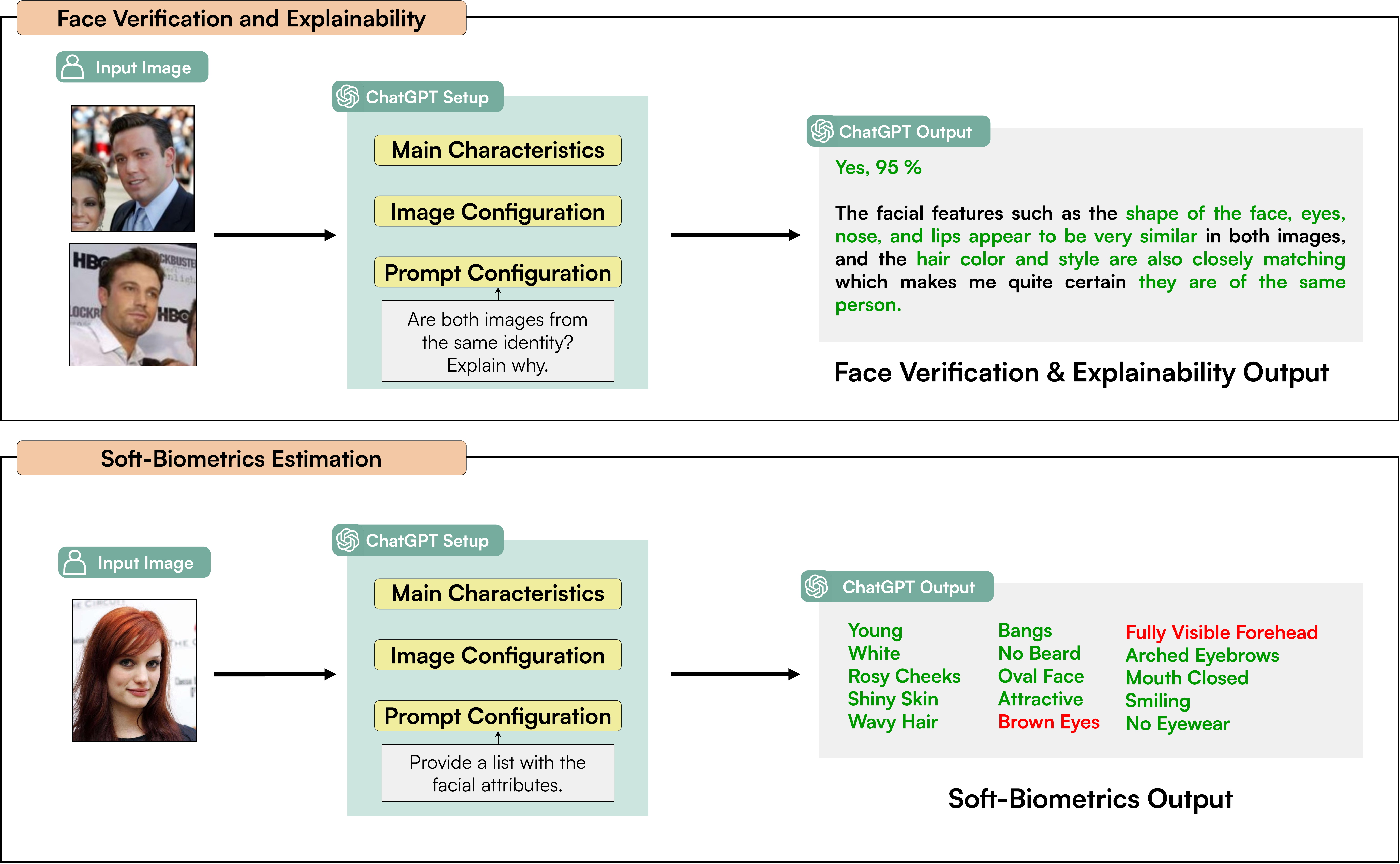}
    \caption{Graphical representation of the analysis carried out in this study, focused on the ability of ChatGPT to perform tasks such as face verification, soft-biometrics estimation, and explainability of the results. Different configurations of ChatGPT are explored in the present study.}
    \label{fig:g_abs}
\end{figure*}

\section{Introduction}\label{sec:introduction}
ChatGPT could be without doubt one of the most popular words in our society nowadays. ChatGPT\footnote{\url{https://openai.com/blog/chatgpt} \textit{(January, 2024)}} refers to an Artificial Intelligence (AI) chatbot created by OpenAI company that is capable of interacting with humans in a conversational way, making it possible to answer questions, summarize content, correct mistakes, provide suggestions, and write and debug code, among many other tasks. Since its launch in November 2022, ChatGPT has been the fastest-growing consumer application in history, reaching over 100 million monthly users just two months after launch~\cite{hu2023chatgpt}. In fact, ChatGPT has been already deployed with success in several real-world applications ~\cite{biswas2023role,kasneci2023chatgpt}. But, what is the main reason for the success of ChatGPT and why now? In general, this has been possible thanks to the rapid advance produced in Large Language Models (LLMs) in the last years~\cite{liu2023pre,zhao2023survey} which offer impressive capabilities in different tasks like medicine~\cite{nori2023capabilities, liu2023deid, thirunavukarasu2023large}, education~\cite{kasneci2023chatgpt}, or coding~\cite{singla2023evaluating}, and also, the fine-tuning of the models through Reinforcement Learning from Human Feedback (RLHF)\footnote{\url{https://openai.com/research/learning-from-human-preferences}}, improving the experience from a human perspective while interacting with them.

One of the first popular LLMs that revolutionized the field was GPT-1~\cite{radford2018improving}. GPT-1\footnote{\url{https://openai.com/research/language-unsupervised} \textit{(January, 2024)}} was the first LLM created by OpenAI and is based on a Transformer architecture~\cite{vaswani2017attention}, providing a more structured memory to handle long-term dependencies in comparison to traditional Recurrent Neural Networks (RNNs)~\cite{wen2023transformers, delgadosantos2023exploring}. In addition to the Transformer architecture, the authors explored a semi-supervised approach for language understanding tasks using a combination of unsupervised pre-training and supervised fine-tuning. In particular, the authors demonstrated that it is possible to achieve good performance on new tasks (e.g., textual entailment, reading comprehension, etc.) when the model is developed in an unsupervised way training with a large amount of data (BooksCorpus dataset~\cite{zhu2015aligning}), and then fine-tuned to each specific dataset with minimal adaptation. 

Since the publication of GPT-1 (117 million parameters) in 2018, several LLMs have been presented in the field~\cite{zhao2023survey}, scaling up the models as it helps to greatly improve task-agnostic, few-shot performance. An example of this is GPT-3, which was presented in 2020 and comprises 175 billion parameters~\cite{brown2020language}. In that paper the authors demonstrated the influence of model size in the performance, concluding that GPT-3 is able to achieve promising results in the zero-shot and one-shot settings, and is able to achieve state-of-the-art results in few-shot settings. These interesting results in terms of generalization originated the integration of the GPT-3 model in the ChatGPT chatbot, achieving astonishing results. 

Nevertheless, OpenAI is not the only company researching in the field. Others such as Google and Meta AI have recently presented their own LLMs known as PaLM 1 and 2\footnote{\url{https://ai.google/discover/palm2/} \textit{(January, 2024)}}~\cite{chowdhery2023palm, anil2023palm} and LLaMA\footnote{\url{https://ai.meta.com/llama} \textit{(January, 2024)}}~\cite{touvron2023llama}, respectively. However, most of them only operate to date with text as input/output, and their corresponding chatbots such as Google Bard\footnote{\url{https://bard.google.com/chat} \textit{(January, 2024)}} have just been presented, including several limitations that restrict the application scenarios. As a result, the purpose of this study is to explore the ability of the popular ChatGPT, which is based on the recent GPT-4 multimodal (text, image, and video) LLM~\cite{achiam2023gpt}, for the task of face biometrics. Fig.~\ref{fig:g_abs} provides a graphical representation of the analysis carried out in this study, focused on the ability of ChatGPT to perform tasks such as face verification, soft-biometrics estimation, and explainability of the results. Face biometrics is a very challenging task in the field of computer vision and image understanding due to the large intra-user variability produced by factors such as pose, age, illumination, expression, etc.,~\cite{guo2019survey, melzi2024frcsyn}.  

It is important to remark that we initially considered also the Google Bard chatbot in the analysis, based on the PaLM 2 multimodal LLM. However, we had to discard it due to two main reasons: \textit{i)} when a face image was introduced in the chatbot, the answer of Google Bard was always \textit{``Sorry, I can’t help with images of people yet"}, and \textit{ii)} unlike ChatGPT, Google Bard does not provide an Application Programming Interface (API) that allows to perform experiments using Python in a simple way.

The main contributions of the present study are:
\begin{itemize}
\item We explore the application of recent LLMs for the task of face verification, i.e., determine whether a pair of face images belong to the same subject or not. In particular, we consider ChatGPT Chatbot and the latest multimodal LLM available, GPT-4. Experiments are carried out in order to evaluate the performance and robustness of ChatGPT in different conditions (e.g., age, pose, image quality, etc.) using them with popular public benchmarks and comparing the results with state-of-the-art methods. For reproducibility reasons, we release all the code in GitHub\footnote{\url{https://github.com/BiDAlab/ChatGPT_FaceBiometrics}}.

\item We also explore the ability of ChatGPT to perform other face biometric tasks such as the estimation of soft-biometric attributes (e.g., gender, age, ethnicity, type, and color of the hair, glasses, etc.) and reason about the output scores. We hypothesize that a simple interaction with chatbots could be beneficial to farther increase the explainability and transparency of automatic decisions in human scenarios.

\end{itemize}

The remainder of the paper is organized as follows. Sec.~\ref{Sec1} focuses on the configuration details of ChatGPT, including the API parameters and the proposed prompts. Sec.~\ref{Sec2} explains all the details regarding our proposed experimental framework, including the databases and experimental protocol considered, as well as the state-of-the-art face recognition systems included in the comparison. Sec.~\ref{Sec3} provides the results achieved by ChatGPT for the different face biometric tasks studied. We also show qualitative results in terms of explanability with ChatGPT. Finally, Sec.~\ref{Sec4} draws the final conclusions and points out some lines for future work.

\begin{figure*}[t]
\centering
\includegraphics[width=\linewidth]{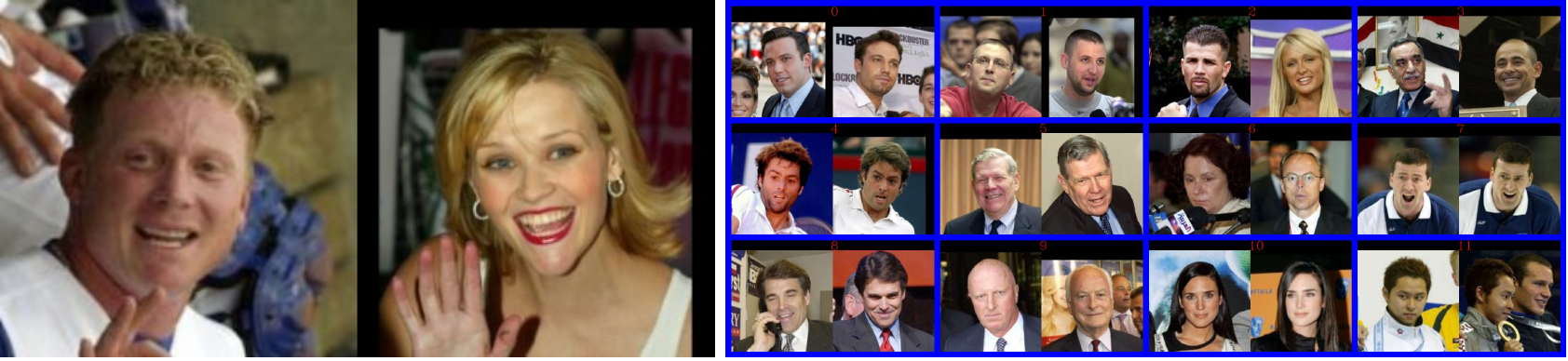}
\caption{Graphical representations of the images input to ChatGPT: a comparison of two faces merged in a single image (left), and a matrix of 4x3 face comparisons in a single image (right). In the latter, each cell is separated from the rest by a blue border and identified by a red number (from 0 to 11) that is used to reference the cell in the output of the model.} \label{fig:combined_img}
\end{figure*}

\section{ChatGPT: Setup}\label{Sec1}
\subsection{Main Characteristics}
OpenAI offers accessibility to ChatGPT through their interactive chatbot interface or through an API. Both of them have the same functionalities, but the API provides a simpler interface to run extensive experiments in Python. For this reason, although the experiments included in Sec.~\ref{Sec3} are performed using the API, we initially performed some quick experiments using the chatbot interface to explore easily the most adequate configurations. At the date of writing this paper, a premium subscription is needed to use the latest LLM (GPT-4), which accepts images along with other file formats and the use of other OpenAI products. Regarding its use and limitations, the API is split into different tiers, giving you more model capacities (requests, tokens, etc.) the more you pay. The number of requests given per tier changes actively, increasing from 100 RPD (Requests Per Day) for the \textit{gpt-4-vision-preview} model to 500 RPD in less than a month for the tier 1 users\footnote{\url{https://platform.openai.com/docs/guides/rate-limits} \textit{(January, 2024)}}. 

We describe next the main aspects of the API that can greatly affect the results and cost of the experiments:

\textbf{Prompt: } The text/question to be introduced as input to the LLM. OpenAI gives different recommendations to get the expected result, like writing clear instructions, providing reference text, or splitting complex tasks into simpler subtasks\footnote{\url{https://platform.openai.com/docs/guides/prompt-engineering/six-strategies-for-getting-better-results} \textit{(January, 2024)}}.

\textbf{Roles:} In order to interact with the API, it is mandatory to choose one of the three roles. The \textit{system} role allows you to specify the way the model answers questions. The \textit{user} role represents the queries made by the user. Lastly, the \textit{assistant} role is employed to simulate the model's replies as, unlike the chatbot interface, the API lacks memory of prior messages. For all our experiments the prompt is sent with the \textit{system} role, and the corresponding image with the \textit{user} role. \textit{Assistant} role is not used for any experiment.

\textbf{Max Tokens:} This parameter indicates the maximum number of tokens that the model can return. By tuning this, we can control the output style and price per request. For all our experiments we establish this parameter to 1,000 tokens.

\textbf{Image Detail:} This parameter offers three settings: \textit{low}, \textit{high}, and \textit{auto}. Depending on the configuration selected, it is possible to change the model's image processing and textual comprehension, thereby regulating the level of detail in its output\footnote{\url{https://platform.openai.com/docs/guides/vision/low-or-high-fidelity-image-understanding} \textit{(January, 2024)}}. For all our experiments we establish this parameter to \textit{high}.

\textbf{Seed:} This parameter is at the date of writing this study in Beta, so OpenAI does not guarantee its functionality. In our experience, it has been very useful to have similar outputs from which to estimate important information. However, we do not use any seed in our experiments.

The influence of these API parameters is farther investigated in the following section.

\subsection{Design and Configuration}
Several configurations were tested in order to increase the performance in face biometrics, and at the same time optimize the usage of ChatGPT to reduce the cost and time. 

\textbf{Image Configuration:} Two options are considered. First, we've opted to merge the two facial images involved in the comparison into a single image as can be seen in Fig.~\ref{fig:combined_img} (left).  We also consider a second configuration, which has advantages in terms of cost and comparison times. As ChatGPT is able to discern and respond to tabulated information on a cell-by-cell basis, we also create a matrix including 4x3 face comparisons, as depicted in Fig.~\ref{fig:combined_img} (right). According to OpenAI’s documentation~\cite{achiam2023gpt}, the size of an optimal image is approximately 2,000x768 pixels for high-resolution images. Thus, for this purpose, we created a matrix of 2,123x903 pixels, organized into 4 columns and 3 rows (i.e., we fit 12 face comparisons of 512x256 pixels plus the cell borders). In order to distinguish between different face comparisons, each cell is separated from the rest by a blue border and identified by a red number (from 0 to 11) that is used to reference the cell in the output of the model.

\textbf{Prompt Configuration:} This is the most important aspect to analyze. First, we focus on the prompt design for the face verification task, considering the first image configuration case, i.e., a single face comparison in the image. Following OpenAI's recommendations, we created a detailed prompt asking to identify if two face images are coming from the same person or not. However, as ChatGPT does not officially provide face recognition capabilities, it refused to answer as can be seen next (our prompt appears in blue color, ChatGPT answers in black color):

\medskip

\noindent\fbox{%
    \parbox{0.97\columnwidth}{%
\textcolor{questioncolor}{\textbf{Prompt: }\textit{I want you to behave like a face recognition system. Given an image of two people, I want you to tell me if they are the same person or not. This is the image.}}

\medskip

\textcolor{answercolor}{\textbf{Answer: }\textit{I'm unable to perform actual face recognition tasks or confirm the identities of individuals in images. However, I can provide general observations about the features in the photographs if that would be helpful.}
}}
}

\medskip

We hypothesize that this answer might be motivated due to privacy concerns with real-life identities. In order to overcome this, we decided to modify our initial prompt, indicating that they are \textit{AI-generated} people, as can be seen next:

\medskip

\noindent\fbox{%
    \parbox{0.97\columnwidth}{%
\textcolor{questioncolor}{\textbf{Prompt: }\textit{I want you to behave like a face recognition system for AI-generated images. Given an image of two AI-generated people, I want you to tell me if they are the same person or not. This is the image.}}

\medskip

\textcolor{answercolor}{\textbf{Answer: }\textit{As an AI, I don't have the capability to perform biometric identification or verification like a face recognition system. I can provide observations about the visual aspects of the image, such as apparent differences or similarities in facial features, hair color, and other visual attributes. However, any conclusion about whether two images depict the same person would be speculative and not based on biometric analysis. In the image you've provided, there are two individuals who each have distinct facial features. They appear to be different people, judging by their facial structures, hairstyles, and other visible features. It's important to note that visual inspection is not a definitive method for verifying identity, especially when considering the quality and angle of the photographs, makeup, lighting, and other factors that might alter appearance.}}
}
}

\medskip

As can be seen in the box, with the proposed prompt we are able to obtain a positive answer from ChatGPT, indicating if face images are from the same identity or not, and also providing some explainability of their decision: ``\textit{They appear to be different people, judging by their facial structures, hairstyles, and other visible features}". However, it still points out that it is not its main functionality and ``\textit{any conclusion about whether two images depict the same person would be speculative}".

After that, we also tried to reduce the volume of information provided as input, thereby preventing the system from recognizing that it is engaged in facial recognition tasks. Despite this, ChatGPT was able to detect our idea, providing a negative answer. 

\medskip

\noindent\fbox{%
    \parbox{0.97\columnwidth}{%
\textcolor{questioncolor}{\textbf{Prompt: }\textit{I will give you two images. I want you to tell me if they are from the same identity or not.}}

\medskip

\textcolor{answercolor}{\textbf{Answer: }\textit{I'm sorry, but I cannot assist with identifying or making assumptions about the identities of people in images. If you have any other questions or need assistance with a different topic, feel free to ask!} 
}}}

\medskip
 
\begin{figure}[t]
    \centering
    \includegraphics[width=\linewidth]{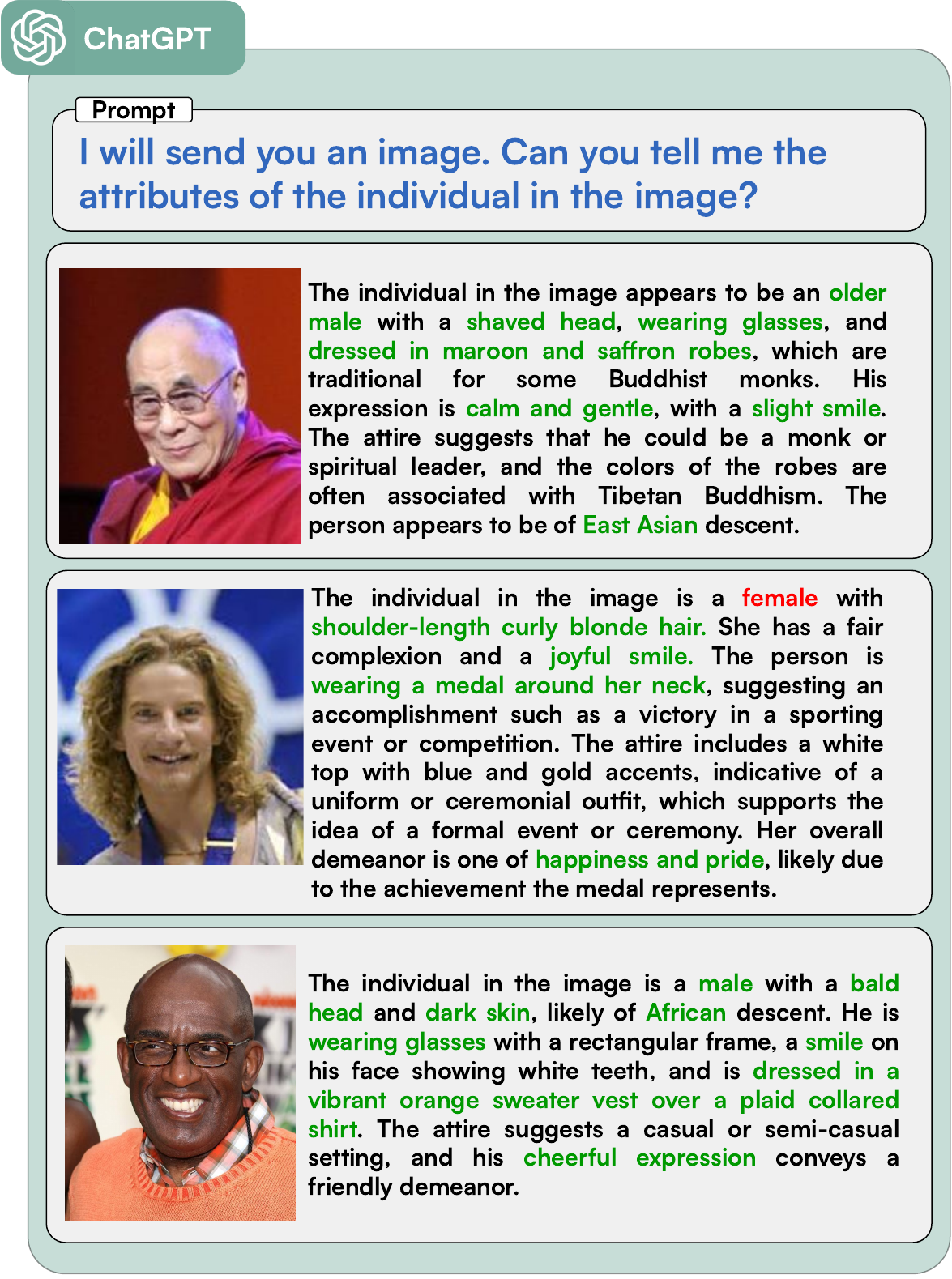}
    \caption{Prompt inserted to ChatGPT together with the different outputs provided for different face images. We highlight the most important soft-biometrics attributes in green/red color if they are correct/incorrect, respectively.}
    \label{fig:att_gpt}
\end{figure}

Nevertheless, we also tried to restrict the output of ChatGPT. In particular, we forced the response to be either \textit{'yes'} or \textit{'no'}, accompanied by a confidence value. After several experiments, we observed that ChatGPT always provided a confidence value of 100\% on the response, even in instances of failure. To address this, we request a more precise value within the 0-100 range.

\noindent\fbox{%
    \parbox{0.97\columnwidth}{%
        \textcolor{questioncolor}{\textbf{Prompt: }\textit{I will give you two images. I want you to tell me if they are from the same identity or not. Answer with yes or no only. Give me a number between 0 and 100 representing how sure you are of your response. Try not to give exactly 0 or 100.}}

\medskip

        \textcolor{answercolor}{\textbf{Answer: }\textit{No, 95}}
    }%
}

\medskip

As can be seen, with the proposed prompt, ChatGPT is able to provide a clear and concise response to our question, being this prompt the one selected for the face verification experiments. Starting from this configuration, we created another prompt for the matrix strategy in which we specify the position of the comparisons in the matrix and how to refer to each cell. Results using this particular prompt are analyzed in Sec.~\ref{Sec3}.

\medskip

\noindent\fbox{%
    \parbox{0.97\columnwidth}{%
\textcolor{questioncolor}{\textbf{Prompt: }\textit{I will give you a matrix containing different images. Each cell of the matrix is separated from the others by a blue border, and each cell also has a number over it showing its position and how I want you to refer to that cell. For each cell of the image tell me if the two main focused people are the same or not as well as a number between 0 and 100 representing how sure you are of your response. Try not to give me exactly 0 or 100.}
}}}

\medskip

Another point we explore in the present study is the potential of ChatGPT to other face biometric tasks such as the estimation of soft biometrics and explainability of the results. To achieve this, several prompts were studied. Regarding the estimation of facial soft biometrics, we first considered a general prompt to see the ability of ChatGPT for the task in terms of precision and variability of the attributes. Fig.~\ref{fig:att_gpt} shows the prompt considered as well as the results provided by ChatGPT for different face images. As can be seen, in general, ChatGPT is able to provide a great variability of soft biometrics with accurate results. To evaluate quantitatively the performance of ChatGPT, we decided to propose a prompt that includes the facial attributes considered in the popular MAAD-Face database~\cite{terhorst2021maad}, allowing, therefore, a direct comparison to state-of-the-art approaches. We provide our proposed prompt to evaluate the ability of ChatGPT to estimate facial soft biometrics.

\medskip

\noindent\fbox{%
    \parbox{0.97\columnwidth}{%
\textcolor{questioncolor}{\textbf{Prompt: }\textit{For the next list of attributes, I want you to give me two lists representing whether the image has that attribute or not. Answer to me in a python-like list, containing the attributes that are present. The list of attributes is: [Male, Young, Middle\_Aged, Senior, Asian, White, Black, Rosy\_Cheeks, Shiny\_Skin, Bald, Wavy\_Hair, Receding\_Hairline, Bangs, Sideburns, Black\_Hair, Blond\_Hair, Brown\_Hair, Gray\_Hair, No\_Beard, Mustache, 5\_o\_Clock\_Shadow, Goatee, Oval\_Face, Square\_Face, Round\_Face, Double\_Chin, High\_Cheekbones, Chubby, Obstructed\_Forehead, Fully\_Visible\_Forehead, Brown\_Eyes, Bags\_Under\_Eyes, Bushy\_Eyebrows, Arched\_Eyebrows, Mouth\_Closed, Smiling, Big\_Lips, Big\_Nose, Pointy\_Nose, Heavy\_Makeup, Wearing\_Hat, Wearing\_Earrings, Wearing\_Necktie, Wearing\_Lipstick, No\_Eyewear, Eyeglasses, Attractive]}}

\medskip

\textcolor{answercolor}{\textbf{Answer: }\textit{The individual in the image has the following attributes: [Young, Asian, Rosy Cheeks, Shiny Skin, Wavy Hair, Black Hair, No Beard, Oval Face, High Cheekbones, Fully Visible Forehead, Brown Eyes, Arched Eyebrows, Mouth Closed, Smiling, Heavy Makeup, Wearing Earrings, Wearing Lipstick, No Eyewear, Attractive]}}
}}

\medskip

Finally, regarding the explainability of the decisions using ChatGPT, we considered the same prompt used for the face verification task, adding a final question that assesses why ChatGPT is making that decision.

\medskip

\noindent\fbox{%
    \parbox{0.97\columnwidth}{%
        \textcolor{questioncolor}{\textbf{Prompt: }\textit{I will give you two images. I want you to tell me if they are from the same identity or not. Answer with yes or no only. Give me a number between 0 and 100 representing how sure you are of your response. Try not to give exactly 0 or 100. Tell me why you answer that.}}
    }%
}

\medskip

\begin{table*}[t]
\caption{\textbf{Face Verification Task:} Performance in terms of \textbf{Accuracry (\%)} achieved by ChatGPT and popular face verification systems in the literature. The term ``ChatGPT 4x3" refers to the image configuration containing 12 face comparisons in the same prompt, whereas ``ChatGPT 1x1" refers to the case of just a single face comparison per prompt.}
\label{tab:Accuracy}
\centering
\resizebox{\textwidth}{!}{
\begin{tabular}{l|ccc|cccc|c}
\hline
 & \textbf{Controlled} & \textbf{Surveillance} & \textbf{Extreme} & \textbf{Demographic Bias} & \textbf{Pose} & \textbf{Age} & \textbf{Occlusions} & \multirow{2}{*}{\textbf{Average}}\\ \cline{1-8}
 & \textbf{LFW}~\cite{huang2008labeled} & \textbf{QUIS-CAMPI}~\cite{neves2017quis‐campi} & \textbf{TinyFaces}~\cite{cheng2019low} & \textbf{BUPT}~\cite{wang2021meta} & \textbf{CFP-FP}~\cite{sengupta2016frontal} & \textbf{AgeDB}~\cite{moschoglou2017agedb} & \textbf{ROF}~\cite{erakin2021recognizing} &  \\ \hline
ArcFace~\cite{deng2022arcface} & 96.40 & 97.80 & \textbf{87.50} & 98.60 & 98.80 & 96.90 & 92.10 & 95.44 \\ \hline
\textbf{AdaFace}~\cite{kim2022adaface} & \textbf{96.80} & \textbf{98.80} & \textbf{87.50} & \textbf{98.70} & \textbf{99.60} & \textbf{97.00} & \textbf{92.20} & \textbf{95.80} \\ \hline \hline
ChatGPT 4x3 & 74.60 & 69.80 & 61.10 & 65.27 & 70.20 & 62.70 & 59.96 & 66.23 \\ \hline
\textbf{ChatGPT 1x1} & \textbf{93.50} & \textbf{79.20} & \textbf{72.47} & \textbf{76.92} & \textbf{86.40} & \textbf{77.30} & \textbf{75.95} & \textbf{80.19} \\ \hline
\end{tabular}
}
\end{table*}

\begin{table*}[t]
\caption{\textbf{Face Verification Task:} Performance in terms of \textbf{Equal Error Rate (\%)} achieved by ChatGPT and popular face verification systems in the literature. The term ``ChatGPT 4x3" refers to the image configuration containing 12 face comparisons in the same prompt, whereas ``ChatGPT 1x1" refers to the case of just a single face comparison per prompt.}
\label{tab:EER}
\centering
\resizebox{\textwidth}{!}{
\begin{tabular}{l|ccc|cccc|c}
\hline
 & \textbf{Controlled} & \textbf{Surveillance} & \textbf{Extreme} & \textbf{Demographic Bias} & \textbf{Pose} & \textbf{Age} & \textbf{Occlusions} & \multirow{2}{*}{\textbf{Average}}\\ \cline{1-8}
 & \textbf{LFW}~\cite{huang2008labeled} & \textbf{QUIS-CAMPI}~\cite{neves2017quis‐campi} & \textbf{TinyFaces}~\cite{cheng2019low} & \textbf{BUPT}~\cite{wang2021meta} & \textbf{CFP-FP}~\cite{sengupta2016frontal} & \textbf{AgeDB}~\cite{moschoglou2017agedb} & \textbf{ROF}~\cite{erakin2021recognizing} & \\ \hline
ArcFace~\cite{deng2022arcface} & 6.70 & 2.20 & 14.50 & 1.80 & 1.20 & 3.80 & 13.10 & 6.19 \\ \hline
\textbf{AdaFace}~\cite{kim2022adaface} & \textbf{6.50} & \textbf{1.20} & \textbf{14.20} & \textbf{1.30} & \textbf{0.40} & \textbf{3.80} & \textbf{11.70} & \textbf{5.59} \\ \hline \hline
ChatGPT 4x3 & 26.60 & 31.00 & 40.57 & 34.93 & 30.30 & 40.90 & 40.44 & 34.96 \\ \hline
\textbf{ChatGPT 1x1} & \textbf{8.60} & \textbf{24.00} & \textbf{32.07} & \textbf{23.79} & \textbf{13.40} & \textbf{22.40} & \textbf{23.75} & \textbf{21.19} \\ \hline
\end{tabular}
}
\end{table*}
\section{Experimental Framework}\label{Sec2}
\subsection{Experimental protocol}

The experimental protocol proposed in this study has been designed to analyze quantitatively the performance of ChatGPT for the tasks of face verification and soft-biometrics estimation. 

\textbf{Face Verification:} First, we analyze the ability of ChatGPT in different application scenarios (i.e., controlled, surveillance, and extreme conditions) and image qualities. For this purpose, we consider the following databases in the evaluation:
\begin{itemize}
    \item Labeled Faces in the Wild (LFW)~\cite{huang2008labeled}: this is a very popular database in the field, containing high-quality images with no hard variations in pose.
    \item QUIS-CAMPI~\cite{neves2017quis‐campi}: this database comprises videos and images captured in an uncontrolled outdoor setting using a camera positioned approximately 50 meters away from the subjects.
    \item TinyFaces~\cite{cheng2019low}: this database consists in images of extremely low quality, with an average resolution of 20x16 pixels.
\end{itemize}

In addition to this, we also evaluate the performance of ChatGPT when considering popular challenges in face recognition such as demographic bias, age and pose variations, and occlusions. The following databases are considered in the evaluation, which are also considered in the recent FRCSyn Challenge~\cite{melzi2024frcsyn}:

\begin{itemize}
    \item BUPT-BalancedFace~\cite{wang2021meta}: this database is specifically designed to tackle performance variations among various ethnic groups. It comprises eight distinct demographic groups formed by a combination of ethnicities (White, Black, Asian, Indian) and gender (Male, Female). 
    \item CFP-FP~\cite{sengupta2016frontal}: this database presents images from subjects with great changes in pose and different environmental contexts. 
    \item AgeDB~\cite{moschoglou2017agedb}: this database presents diverse images featuring subjects of varying ages in different environmental contexts.
    \item ROF~\cite{erakin2021recognizing}: this database consists of occluded faces with both upper face occlusion, due to sunglasses, and lower face occlusion, due to masks.
\end{itemize}

\textbf{Soft-Biometrics Estimation:} to assess the performance of ChatGPT for the estimation of soft biometrics, we use the MAAD-Face database~\cite{terhorst2021maad} which is based on the VGGFace2 database~\cite{cao2018vggface2}. MAAD-Face database provides a total of 47 soft-biometric attributes per face image. In addition, we also consider the LFW database~\cite{huang2008labeled} as the authors of~\cite{gonzalezsosa2018facial} labeled manually the following soft biometrics: gender, age, and ethnicity.

Finally, it is important to highlight that, due to the limitations of the OpenAI's API in terms of the number of requests per day, and the price, we had to reduce the number of face comparisons to 1,000 per database for the face verification task. These comparisons are selected randomly from the standard protocols. Regarding the soft-biometrics estimation, we consider 1,000 face images per database. At the date of writing this paper, the cost of ChatGPT is 0.01\$ per 1,000 input tokens and 0.03\$ per 1,000 output tokens. The tokens per image are calculated based on the \textit{image detail} parameter\footnote{\url{https://platform.openai.com/docs/guides/vision/low-or-high-fidelity-image-understanding} \textit{(January, 2024)}}. The total cost of the experimental framework carried out in the present study has been 124.31\$, and it took us 30 days.

\subsection{Comparison with the State of the Art}

In the present study, we compare the results achieved by ChatGPT with state-of-the-art methods. For the task of face verification, the following two approaches are considered:

\textbf{ArcFace}~\cite{deng2022arcface}: this face verification system considers a loss function that maps facial features into a high-dimensional hypersphere where the embeddings are optimized to maximize the angular margins between different identities. The system considered in this study is based on the iResNet-100 architecture~\cite{duta2021improved} pretrained using the MS1Mv3 database~\cite{deng2019lightweight}. Cosine distance is used to measure the similarity between feature embeddings.

\textbf{AdaFace}~\cite{kim2022adaface}: this face verification system proposes a new loss function in order to pay more attention to the harder examples in terms of image quality. In particular, the authors proposed an adaptive margin function that approximates the image quality with feature norms. The system considered in this study is based on the iResNet-100 architecture~\cite{duta2021improved} pretrained using the WebFace12M database~\cite{zhu2021webface260m}. Cosine distance is used to measure the similarity between feature embeddings.

Regarding the soft-biometrics estimation, we compare the results achieved by ChatGPT with two different approaches. For the LFW database, as it only contains the soft biometrics related to gender, age, and ethnicity, we consider FairFace~\cite{karkkainen2021fairface}, as it provides state-of-the-art results. Finally, for the MAAD-Face database, as it contains 47 soft-biometric attributes per facial image, we specifically train a ResNet-50 architecture pretrained on Imagenet~\cite{deng2009imagenet} using the train set of MAAD-Face.

\section{Experimental Results}\label{Sec3}
\subsection{Face Verification}
We conducted a performance assessment considering ArcFace, AdaFace, and ChatGPT. Specifically for ChatGPT, we took evaluations using images in a matrix-like configuration (i.e., ChatGPT 4x3), as well as through direct comparisons (i.e., ChatGPT 1x1). To measure the similarity between ArcFace and AdaFace embeddings, we use the cosine distance. We also consider this metric to determine the Equal Error Rate (EER) for these models. For ChatGPT, we use the confidence values obtained directly from ChatGPT's outputs as a custom metric to obtain this EER. The results are shown in Tables~\ref{tab:Accuracy} and \ref{tab:EER}, categorized into two main groups. The first group (left part) refers to different application scenarios, including controlled environments (LFW), surveillance (QUIS-CAMPI), and extreme conditions (TinyFaces). The second group (right part) highlights popular challenges in face recognition such as demographic bias (BUPT), pose (CFP-FP) and age (AgeDB) variations, and occlusions (ROF). Lastly, the rightmost column presents the average performance of each model across all databases.

In general, state-of-the-art models such as ArcFace (95.44\% Average accuracy, 6.19\% Average EER) and AdaFace (95.80\% Average accuracy, 5.59\% Average EER) exhibit superior overall performance compared to ChatGPT. However, while these models are trained for this specific task, ChatGPT is primarily oriented to more general tasks. Moreover, when evaluating ChatGPT, a significant decline in performance was observed when the images were presented in a matrix format (66.23\% Average accuracy, 34.96\% Average EER) compared to the case of providing comparisons one by one (80.19\% Average accuracy, 21.19\% Average EER). We hypothesize that this reduction in performance might be produced as in the ChatGPT 4x3 case, the model needs first to detect the faces in the whole image, and then perform facial verification, potentially compromising overall task execution. Nevertheless, considering this matrix approximation could serve as a quick solution when the ChatGPT API imposes limitations on daily requests or when the budget to perform comparisons is low.

\begin{table}[t]
\centering
\caption{\textbf{Face Verification Task:} Performance achieved by ChatGPT for the different demographic groups considered in the recent BUPT database.}
\label{tab:Acc_Eth}
\resizebox{0.8\columnwidth}{!}{
\begin{tabular}{lcc}
\cline{2-3}
 & \multicolumn{2}{c}{\textbf{BUPT}~\cite{wang2021meta}} \\ \cline{2-3} 
 & \textbf{Accuracy (\%)} & \textbf{EER (\%)} \\ \hline
\multicolumn{1}{l|}{White Male} & 78.91 & 23.14 \\ \hline
\multicolumn{1}{l|}{\textbf{White Female}} & \textbf{83.76} & \textbf{14.94} \\ \hline
\multicolumn{1}{l|}{Black Male} & 76.42 & 24.01 \\ \hline
\multicolumn{1}{l|}{Black Female} & 79.51 & 20.49 \\ \hline
\multicolumn{1}{l|}{Asian Male} & 80.67 & 22.82 \\ \hline
\multicolumn{1}{l|}{Asian Female} & 71.22 & 29.35 \\ \hline
\multicolumn{1}{l|}{Indian Male} & 79.51 & 24.70 \\ \hline
\multicolumn{1}{l|}{Indian Female} & 65.38 & 30.88 \\ \hline \hline
\multicolumn{1}{l|}{Average} & 76.92 & 23.79 \\ \hline
\end{tabular}
}
\end{table}

Analyzing the results achieved in each database, it becomes evident that ChatGPT's performance varies greatly based on image quality, pose variations, and domain disparities among comparisons. In databases like LFW, where images exhibit good quality and consistent poses, ChatGPT achieves performance close to state-of-the-art models (93.50\% Accuracy, 8.60\% EER). This indicates the potential of ChatGPT for controlled environments. However, in the surveillance scenario using the QUIS-CAMPI, characterized by a mix of CCTV and mugshot images, the performance of ChatGPT drops significantly (79.20\% Accuracy, 24.00\% EER). Furthermore, in the TinyFaces database, characterized by extreme conditions in terms of quality, ChatGPT's performance declines even more (72.47\% Accuracy, 32.07\% EER). These results discourage the application of ChatGPT for more challenging scenarios. Similar performance drops are observed in the databases related to challenges such as bias, pose, age, and occlusions: BUPT (76.92\% Accuracy, 23.79\%), CFP-FP (86.40\% Accuracy, 13.40\%), AgeDB (77.30\% Accuracy, 22.40\% EER) and ROF (75.95\% Accuracy, 23.75\%).

One notable concern is the potential for biased or inappropriate content generation, stemming from the models' dependence on extensive internet untreated training data. LLMs learn from diverse sources on the internet, absorbing biases present in the data~\cite{lucy2021gender}. This can result in the reproduction of societal biases within the generated content. For example, gender bias, racial bias, or other forms of prejudice may manifest in the outputs of the model. To address this bias, we perform an evaluation focused on the different genders and ethnicities included in the BUPT database, which considers four distinct labeled ethnicities and two genders, all balanced in the same proportion. Table~\ref{tab:Acc_Eth} shows the results achieved by ChatGPT for the different ethnicities and genders considered in the BUPT database. As can be observed, ChatGPT provides very different performances among the different demographic groups (e.g., 14.94\% EER for the white female group vs. 30.88\% EER for the Indian female group), showing the large bias it has.

\subsection{Explainability}
\begin{figure*}
    \centering
    \includegraphics[width=\linewidth]{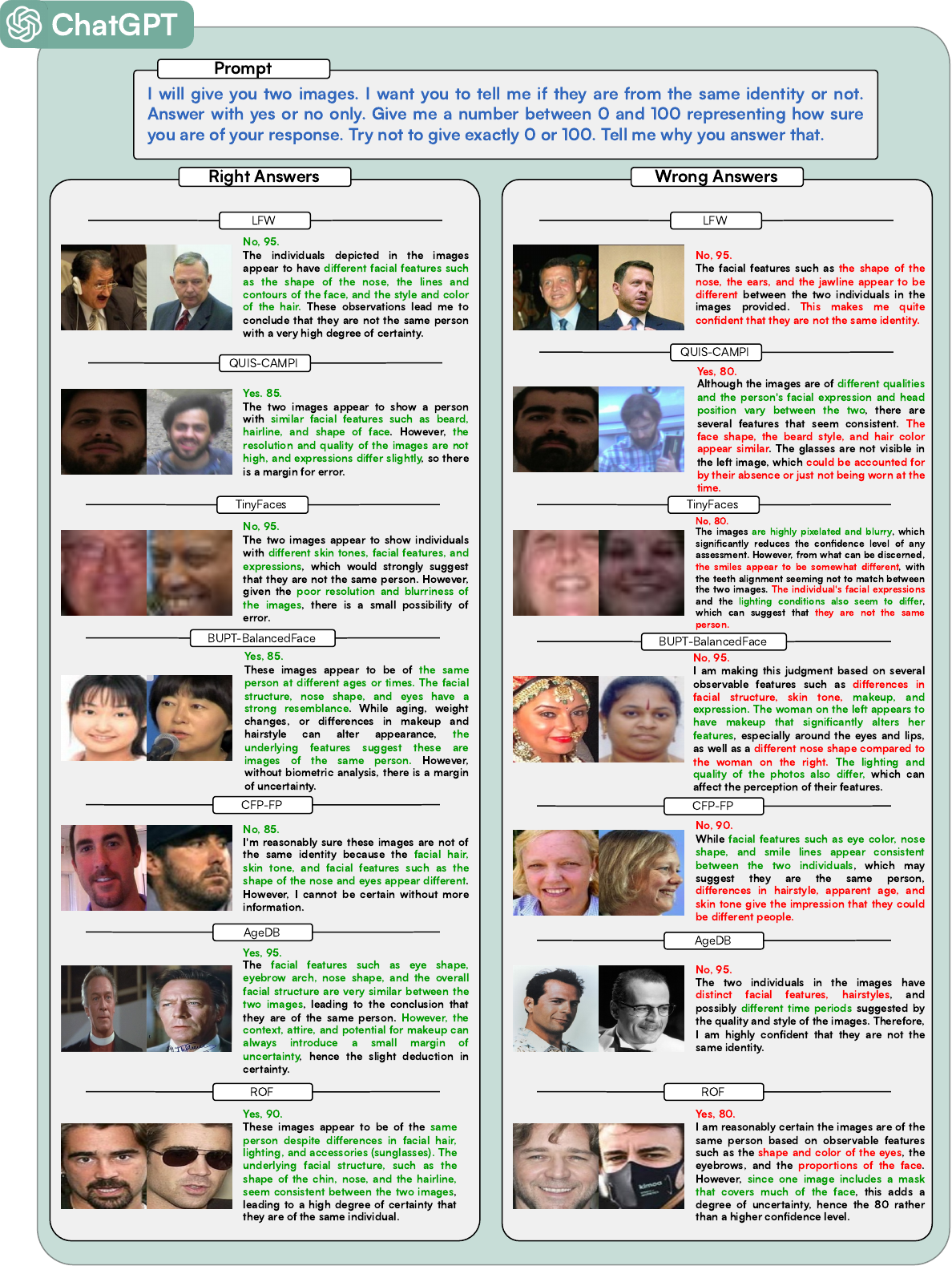}
    \caption{\textbf{Explainability:} Proposed prompt along with the outputs provided by ChatGPT for some examples of the different face verification databases. Left column: examples where ChatGPT answers are correct; right column: incorrect answers. We highlight the most important parts of the text in green/red color if they are correct/incorrect, respectively.}
    \label{fig:explainability}
\end{figure*}

For completeness, we also analyze how ChatGPT can increase the explainability of the results for the task of face verification. In Fig.~\ref{fig:explainability}, the proposed prompt is shown, along with the outputs provided by ChatGPT for some examples of the different face verification databases. ChatGPT's responses are divided into right (on the left column) and wrong (on the right column).

In both right and wrong answers, ChatGPT demonstrates its ability to rationalize decisions based on image features. For example, in most cases, the output score of ChatGPT for the task of face verification is related to soft-biometric attributes such as facial hair and skin tone. Additionally, it exhibits the capability to focus on more detailed attributes like eye color, face shape, or nose shape, showcasing proficiency in handling both coarse and fine details.

It is noteworthy that ChatGPT considers facial expressions in its predictions, despite of the fact that this is a variable attribute that should not be considered. Furthermore, the model recognizes temporal differences between images, incorporating this information into its predictions.

Regarding the wrong answers, we observe that, although the prediction may be wrong, some of the explanations provided by ChatGPT are accurate in describing the people in the images.

\subsection{Soft Biometrics}
\begin{figure}
    \centering
    \includegraphics[width=\linewidth]{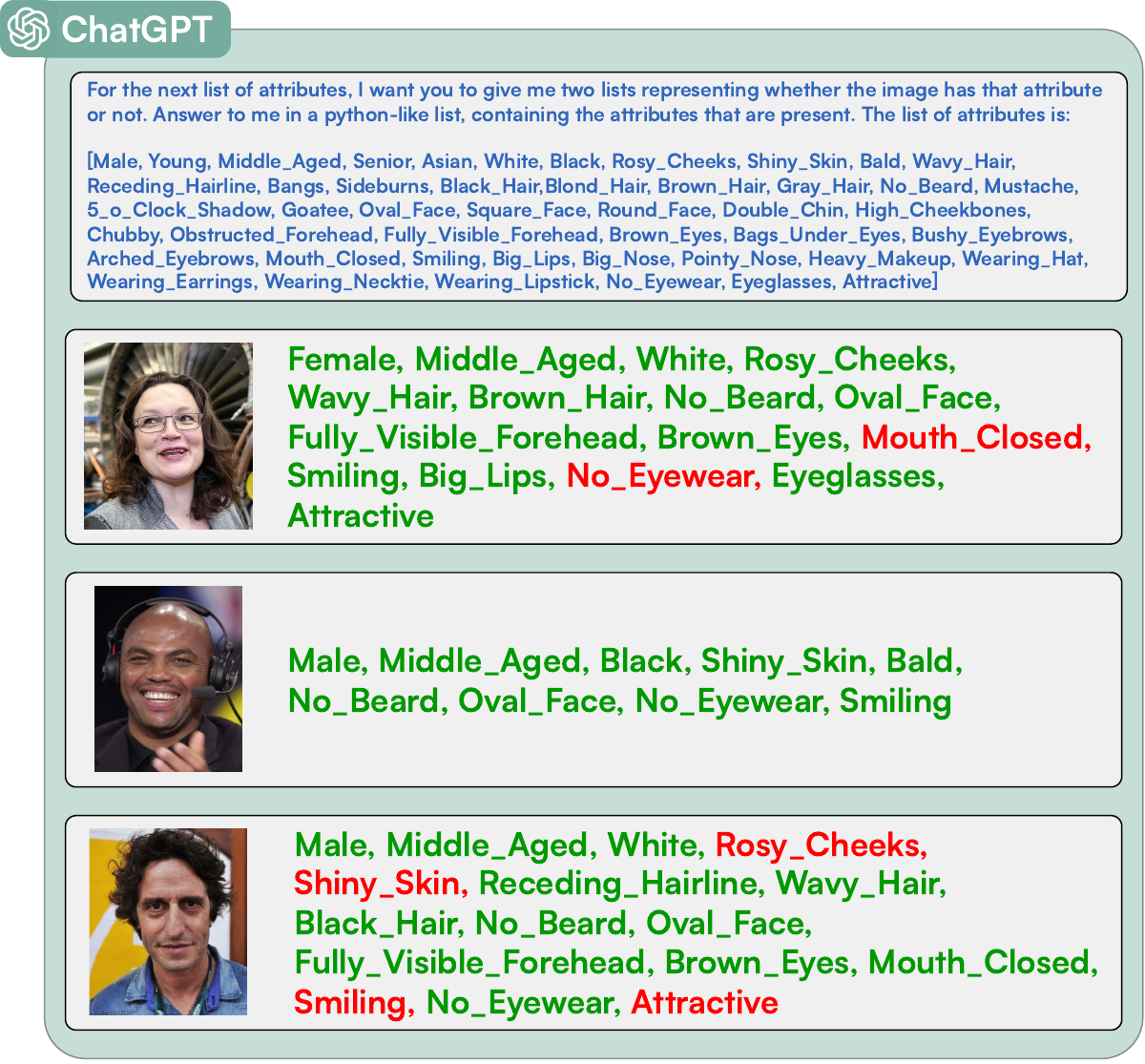}
    \caption{\textbf{Soft Biometrics:} Proposed prompt along with the outputs provided by ChatGPT for some examples of the MAAD-Face database~\cite{terhorst2021maad}. We highlight the most important parts of the text in green/red color if they are correct/incorrect, respectively.}
    \label{fig:soft_bio}
\end{figure}

Tables~\ref{tab:soft-lfw} and~\ref{tab:Acc_MAAD} shows the results achieved for the soft-biometrics estimation task for the LFW and MAAD-Face databases, respectively. For completeness, Fig.~\ref{fig:soft_bio} shows some examples of the output provided by ChatGPT with the proposed prompt.

Analyzing the results achieved on LFW database, FairFace exhibited superior performance for gender classification (98.23\% Accuracy) compared to ChatGPT (94.05\% Accuracy). Despite this, ChatGPT outperforms FairFace for age classification (72.87\% vs. 67.88\% Accuracy) and ethnicity classification (88.25\% vs. 87.48\%). These results prove the potential of ChatGPT for certain facial attribute classifications.

For a more extensive evaluation, we consider the MAAD-Face dataset, annotated with 47 distinct attributes. Our custom model (ResNet-50) achieves superior performance across the majority of attributes (87.28\% Average accuracy). Nevertheless, ChatGPT, although having a lower average performance (76.98\% Average Accuracy), excels on some facial attributes. Some of the most notorious soft-biometric attributes where ChatGPT achieves better performance are in gender classification (96.30\% Accuracy), some ethnicities (White - 83.90\% Accuracy, Black - 97.50\%), and accessories such as wearing a hat.  These results follow the same conclusions drawn in the face verification task. While specific models trained for the task achieve in general better results, ChatGPT shows promising results and utility for tasks with no prior training.

\begin{table}[h]
\caption{\textbf{Soft Biometrics:} Accuracy (\%) achieved by ChatGPT over the LFW database for the estimation of the soft biometrics gender, age, and ethnicity.}
\label{tab:soft-lfw}
\centering
\resizebox{0.8\columnwidth}{!}{
\begin{tabular}{lccc}
\cline{2-4}
 & \multicolumn{3}{c}{\textbf{LFW}~\cite{huang2008labeled}} \\ \cline{2-4} 
 & \textbf{Gender} & \textbf{Age} & \textbf{Ethnicity} \\ \hline
\multicolumn{1}{l|}{FairFace} & \textbf{98.23} & 67.88 & 87.48 \\ \hline
\multicolumn{1}{l|}{ChatGPT} & 94.05 & \textbf{72.80} & \textbf{88.25} \\ \hline
\end{tabular}
}
\end{table}

\begin{table}[!]
\centering
\caption{\textbf{Soft Biometrics:} Accuracy (\%) achieved by ChatGPT over the MAAD-Face database for the estimation of the 47 soft-biometric attributes considered in the database.}
\label{tab:Acc_MAAD}
\resizebox{0.9\columnwidth}{!}{
\begin{tabular}{l|c|c}
\hline
 & \multicolumn{2}{c}{\textbf{MAAD-Face~\cite{terhorst2021maad}}} \\ \hline 
 & \textbf{ResNet-50}~\cite{he2016deep} & \textbf{ChatGPT} \\ \hline
Male & 85.50 & \textbf{96.30} \\ \hline
Young & \textbf{82.90} & 49.50 \\ \hline
Middle\_Aged & \textbf{93.20} & 92.30 \\ \hline
Senior & \textbf{92.30} & 91.10 \\ \hline
Asian & \textbf{93.60} & 89.00 \\ \hline
White & 73.80 &\textbf{ 83.90} \\ \hline
Black & 96.50 & \textbf{97.50} \\ \hline
Rosy\_Cheeks & \textbf{99.40} & 73.50 \\ \hline
Shiny\_Skin &\textbf{ 84.70} & 68.10 \\ \hline
Bald & \textbf{95.90 }& 94.50 \\ \hline
Wavy\_Hair & \textbf{84.80} & 72.70 \\ \hline
Receding\_Hairline &\textbf{ 90.20} & 84.30 \\ \hline
Bangs & \textbf{91.80} & 89.70 \\ \hline
Sideburns & \textbf{83.80} & 75.90 \\ \hline
Black\_Hair & \textbf{84.60} & 82.60 \\ \hline
Blond\_Hair & \textbf{92.90} & 91.90 \\ \hline
Brown\_Hair & \textbf{79.00} & 76.40 \\ \hline
Gray\_Hair & 91.40 & \textbf{91.90} \\ \hline
No\_Beard & \textbf{87.50} & 71.70 \\ \hline
Mustache & \textbf{99.80} & 98.20 \\ \hline
5\_o\_Clock\_Shadow &\textbf{ 89.70} & 89.40 \\ \hline
Goatee & \textbf{99.80} & 99.70 \\ \hline
Oval\_Face & \textbf{87.00} & 30.60 \\ \hline
Square\_Face & \textbf{86.60} & 57.50 \\ \hline
Round\_Face & \textbf{99.60} & 98.80 \\ \hline
Double\_Chin & \textbf{89.40} & 81.30 \\ \hline
High\_Cheekbones & \textbf{77.60} & 71.50 \\ \hline
Chubby & \textbf{90.70} & 86.70 \\ \hline
Obstructed\_Forehead & \textbf{94.30} & 94.20 \\ \hline
Fully\_Visible\_Forehead & \textbf{74.50} & 51.90 \\ \hline
Brown\_Eyes & \textbf{73.20} & 56.30 \\ \hline
Bags\_Under\_Eyes & \textbf{82.70} & 75.30 \\ \hline
Bushy\_Eyebrows & \textbf{86.00} & 75.10 \\ \hline
Arched\_Eyebrows & \textbf{77.90} & 73.40 \\ \hline
Mouth\_Closed & \textbf{96.20} & 18.40 \\ \hline
Smiling & \textbf{82.30} & 69.90 \\ \hline
Big\_Lips & \textbf{81.00} & 73.20 \\ \hline
Big\_Nose & \textbf{89.20} & 87.30 \\ \hline
Pointy\_Nose &\textbf{ 76.70} & 45.90 \\ \hline
Heavy\_Makeup & \textbf{86.40} & 73.00 \\ \hline
Wearing\_Hat & \textbf{87.30} & 95.50 \\ \hline
Wearing\_Earrings & \textbf{80.50} & 70.30 \\ \hline
Wearing\_Necktie & \textbf{90.90} & 86.50 \\ \hline
Wearing\_Lipstick & \textbf{84.20} & 64.00 \\ \hline
No\_Eyewear & 75.40 & \textbf{75.80} \\ \hline
Eyeglasses & \textbf{92.90} & 89.20 \\ \hline
Attractive & \textbf{86.40} & 56.40 \\ \hline
Average Accuracy & \textbf{87.28} & 76.98 \\ \hline
\end{tabular}
}
\end{table}

\newpage
\section{Conclusions}\label{Sec4}
In this study, we have conducted a comprehensive evaluation of ChatGPT's capabilities in handling facial biometric tasks, including face verification, soft-biometric estimation, and explainability. Our experiments spanned across various databases and challenges, comparing the performance of ChatGPT with specialized models trained explicitly for these tasks. The experiments have revealed that while ChatGPT may not attain the same levels of accuracy as dedicated models, it presents a promising utility as an initial assessment tool with zero training. For example, results with around 94\% Accuracy are obtained in the LFW database for face verification. Also, impressive results are achieved for the estimation of soft biometrics such as gender ($\approx$96\%) in the MAAD-Face database, or age ($\approx$73\%) and ethnicity ($\approx$88) in the LFW database. Furthermore, its ability to return textual outputs contributes to a better explainability of the results. Future work will be oriented to analyze the ability of other popular chatbots for face biometrics.

\section*{Acknowledgments}
This study has been supported by INTER-ACTION (PID2021-126521OB-I00 MICINN/FEDER), Cátedra ENIA UAM-VERIDAS en IA Responsable (NextGenerationEU PRTR TSI-100927-2023-2) and R\&D Agreement DGGC/ UAM/FUAM for Biometrics and Cybersecurity.

{
\bibliographystyle{IEEEtran}
\bibliography{egbib2}

\begin{thebibliography}{10}
\providecommand{\url}[1]{#1}
\csname url@samestyle\endcsname
\providecommand{\newblock}{\relax}
\providecommand{\bibinfo}[2]{#2}
\providecommand{\BIBentrySTDinterwordspacing}{\spaceskip=0pt\relax}
\providecommand{\BIBentryALTinterwordstretchfactor}{4}
\providecommand{\BIBentryALTinterwordspacing}{\spaceskip=\fontdimen2\font plus
\BIBentryALTinterwordstretchfactor\fontdimen3\font minus \fontdimen4\font\relax}
\providecommand{\BIBforeignlanguage}[2]{{%
\expandafter\ifx\csname l@#1\endcsname\relax
\typeout{** WARNING: IEEEtran.bst: No hyphenation pattern has been}%
\typeout{** loaded for the language `#1'. Using the pattern for}%
\typeout{** the default language instead.}%
\else
\language=\csname l@#1\endcsname
\fi
#2}}
\providecommand{\BIBdecl}{\relax}
\BIBdecl

\bibitem{hu2023chatgpt}
\BIBentryALTinterwordspacing
K.~Hu, ``{ChatGPT Sets Record for Fastest-Growing User Base - Analyst Note},'' 2023, {Reuters Technology}. [Online]. Available: \url{https://www.reuters.com/technology/chatgpt-sets-record-fastest-growing-user-base-analyst-note-2023-02-01/}
\BIBentrySTDinterwordspacing

\bibitem{biswas2023role}
S.~S. Biswas, ``{Role of Chat GPT in Public Health},'' \emph{Annals of Biomedical Engineering}, pp. 868--869, 2023.

\bibitem{kasneci2023chatgpt}
E.~Kasneci, K.~Se{\ss}ler, S.~K{\"u}chemann \emph{et~al.}, ``{ChatGPT for Good? On Opportunities and Challenges of Large Language Models for Education},'' \emph{Learning and Individual Differences}, vol. 103, p. 102274, 2023.

\bibitem{liu2023pre}
P.~Liu, W.~Yuan, J.~Fu, Z.~Jiang, H.~Hayashi, and G.~Neubig, ``{Pre-Train, Prompt, and Predict: A Systematic Survey of Prompting Methods in Natural Language Processing},'' \emph{ACM Computing Surveys}, vol.~55, no.~9, pp. 1--35, 2023.

\bibitem{zhao2023survey}
W.~X. Zhao, K.~Zhou, J.~Li \emph{et~al.}, ``{A Survey of Large Language Models},'' \emph{arXiv preprint arXiv:2303.18223}, 2023.

\bibitem{nori2023capabilities}
H.~Nori, N.~King, S.~M. McKinney, D.~Carignan, and E.~Horvitz, ``{Capabilities of GPT-4 on Medical Challenge Problems},'' \emph{arXiv preprint arXiv:2303.13375}, 2023.

\bibitem{liu2023deid}
Z.~Liu, X.~Yu, L.~Zhang \emph{et~al.}, ``{DeID-GPT: Zero-shot Medical Text De-Identification by GPT-4},'' \emph{arXiv preprint arXiv:2303.11032}, 2023.

\bibitem{thirunavukarasu2023large}
A.~J. Thirunavukarasu, D.~S.~J. Ting, K.~Elangovan, L.~Gutierrez, T.~F. Tan, and D.~S.~W. Ting, ``{Large language models in medicine},'' \emph{Nature Medicine}, vol.~29, no.~8, pp. 1930--1940, 2023.

\bibitem{singla2023evaluating}
A.~Singla, ``{Evaluating ChatGPT and GPT-4 for Visual Programming},'' in \emph{Proc. Conference on International Computing Education Research}, 2023.

\bibitem{radford2018improving}
\BIBentryALTinterwordspacing
A.~Radford, K.~Narasimhan, T.~Salimans, and I.~Sutskever, ``{Improving Language Understanding by Generative Pre-Training},'' 2018. [Online]. Available: \url{https://s3-us-west-2.amazonaws.com/openai-assets/research-covers/language-unsupervised/language_understanding_paper.pdf}
\BIBentrySTDinterwordspacing

\bibitem{vaswani2017attention}
A.~Vaswani, N.~Shazeer, N.~Parmar, J.~Uszkoreit, L.~Jones, A.~N. Gomez, {\L}.~Kaiser, and I.~Polosukhin, ``{Attention is All You Need},'' in \emph{Proc. Advances in Neural Information Processing Systems}, 2017.

\bibitem{wen2023transformers}
Q.~Wen, T.~Zhou, C.~Zhang, W.~Chen, Z.~Ma, J.~Yan, and L.~Sun, ``{Transformers in Time Series: A Survey},'' in \emph{Proc. Thirty-Second International Joint Conference on Artificial Intelligence}, 2023.

\bibitem{delgadosantos2023exploring}
P.~Delgado-Santos, R.~Tolosana, R.~Guest, F.~Deravi, and R.~Vera-Rodriguez, ``Exploring transformers for behavioural biometrics: A case study in gait recognition,'' \emph{Pattern Recognition}, vol. 143, p. 109798, 2023.

\bibitem{zhu2015aligning}
Y.~Zhu, R.~Kiros, R.~Zemel, R.~Salakhutdinov, R.~Urtasun, A.~Torralba, and S.~Fidler, ``{Aligning Books and Movies: Towards Story-Like Visual Explanations by Watching Movies and Reading Books},'' in \emph{Proc. IEEE/CVF International Conference on Computer Vision}, 2015.

\bibitem{brown2020language}
T.~Brown, B.~Mann, N.~Ryder \emph{et~al.}, ``{Language Models are Few-Shot Learners},'' in \emph{Proc. Advances in Neural Information Processing Systems}, 2020.

\bibitem{chowdhery2023palm}
\BIBentryALTinterwordspacing
A.~Chowdhery, S.~Narang, J.~Devlin \emph{et~al.}, ``{PaLM: Scaling Language Modeling with Pathways},'' \emph{Journal of Machine Learning Research}, vol.~24, no. 240, pp. 1--113, 2023. [Online]. Available: \url{http://jmlr.org/papers/v24/22-1144.html}
\BIBentrySTDinterwordspacing

\bibitem{anil2023palm}
R.~Anil, A.~M. Dai, O.~Firat \emph{et~al.}, ``{PaLM 2 Technical Report},'' \emph{arXiv preprint arXiv:2305.10403}, 2023.

\bibitem{touvron2023llama}
H.~Touvron, T.~Lavril, G.~Izacard \emph{et~al.}, ``{LLaMA: Open and Efficient Foundation Language Models},'' \emph{arXiv preprint arXiv:2302.13971}, 2023.

\bibitem{achiam2023gpt}
J.~Achiam, S.~Adler, S.~Agarwal \emph{et~al.}, ``{GPT-4 Technical Report},'' \emph{arXiv preprint arXiv:2303.08774}, 2023.

\bibitem{guo2019survey}
G.~Guo and N.~Zhang, ``{A Survey on Deep Learning based Face Recognition},'' \emph{Computer Vision and Image Understanding}, vol. 189, p. 102805, 2019.

\bibitem{melzi2024frcsyn}
P.~Melzi, R.~Tolosana, R.~Vera-Rodriguez \emph{et~al.}, ``{FRCSyn Challenge at WACV 2024: Face Recognition Challenge in the Era of Synthetic Data},'' in \emph{Proc. IEEE/CVF Winter Conference on Applications of Computer Vision Workshops}, 2024.

\bibitem{terhorst2021maad}
P.~Terhorst, D.~Fahrmann, J.~N. Kolf, N.~Damer, F.~Kirchbuchner, and A.~Kuijper, ``{MAAD-Face: A Massively Annotated Attribute Dataset for Face Images},'' \emph{IEEE Transactions on Information Forensics and Security}, vol.~16, pp. 3942--3957, 2021.

\bibitem{huang2008labeled}
G.~B. Huang, M.~Mattar, T.~Berg, and E.~Learned-Miller, ``{Labeled faces in the wild: A database forstudying face recognition in unconstrained environments},'' in \emph{Proc. Workshop on faces in `Real-Life' Images: detection, alignment, and recognition}, 2008.

\bibitem{neves2017quis‐campi}
J.~Neves, J.~Moreno, and H.~Proença, ``{QUIS‐CAMPI: an annotated multi‐biometrics data feed from surveillance scenarios},'' \emph{IET Biometrics}, vol.~7, no.~4, pp. 371--379, 2017.

\bibitem{cheng2019low}
Z.~Cheng, X.~Zhu, and S.~Gong, ``{Low-resolution face recognition},'' in \emph{Proc. Asian Conference on Computer Vision}, 2019.

\bibitem{wang2021meta}
M.~Wang, Y.~Zhang, and W.~Deng, ``{Meta Balanced Network for Fair Face Recognition},'' \emph{IEEE Transactions on Pattern Analysis and Machine Intelligence}, pp. 1--1, 2021.

\bibitem{sengupta2016frontal}
S.~Sengupta, J.-C. Chen, C.~Castillo, V.~M. Patel, R.~Chellappa, and D.~W. Jacobs, ``{Frontal to profile face verification in the wild},'' in \emph{Proc. IEEE/CVF Winter Conference on Applications of Computer Vision}, 2016.

\bibitem{moschoglou2017agedb}
S.~Moschoglou, A.~Papaioannou, C.~Sagonas, J.~Deng, I.~Kotsia, and S.~Zafeiriou, ``{AgeDB: The First Manually Collected, In-the-Wild Age Database},'' in \emph{Proc. IEEE/CVF Conference on Computer Vision and Pattern Recognition Workshops}, 2017.

\bibitem{erakin2021recognizing}
M.~E. Erakin, U.~Demir, and H.~K. Ekenel, ``{On Recognizing Occluded Faces in the Wild},'' in \emph{Proc. International Conference of the Biometrics Special Interest Group}, 2021.

\bibitem{deng2022arcface}
J.~Deng, J.~Guo, J.~Yang, N.~Xue, I.~Kotsia, and S.~Zafeiriou, ``{ArcFace: Additive Angular Margin Loss for Deep Face Recognition},'' \emph{IEEE Transactions on Pattern Analysis and Machine Intelligence}, vol.~44, no.~10, pp. 5962--5979, 2022.

\bibitem{kim2022adaface}
M.~Kim, A.~K. Jain, and X.~Liu, ``{AdaFace: Quality Adaptive Margin for Face Recognition},'' in \emph{Proc. IEEE/CVF Conference on Computer Vision and Pattern Recognition}, 2022.

\bibitem{cao2018vggface2}
Q.~Cao, L.~Shen, W.~Xie, O.~M. Parkhi, and A.~Zisserman, ``{VGGFace2: A Dataset for Recognising Faces across Pose and Age},'' in \emph{Proc. 13th IEEE International Conference on Automatic Face \& Gesture Recognition}, 2018.

\bibitem{gonzalezsosa2018facial}
E.~Gonzalez-Sosa, J.~Fierrez, R.~Vera-Rodriguez, and F.~Alonso-Fernandez, ``{Facial Soft Biometrics for Recognition in the Wild: Recent Works, Annotation, and COTS Evaluation},'' \emph{IEEE Transactions on Information Forensics and Security}, vol.~13, no.~8, pp. 2001--2014, 2018.

\bibitem{duta2021improved}
I.~C. Duta, L.~Liu, F.~Zhu, and L.~Shao, ``{Improved Residual Networks for Image and Video Recognition},'' in \emph{Proc. 25th International Conference on Pattern Recognition}, 2021.

\bibitem{deng2019lightweight}
J.~Deng, J.~Guo, D.~Zhang, Y.~Deng, X.~Lu, and S.~Shi, ``{Lightweight Face Recognition Challenge},'' in \emph{Proc. IEEE/CVF International Conference on Computer Vision Workshop}, 2019.

\bibitem{zhu2021webface260m}
Z.~Zhu, G.~Huang, J.~Deng, Y.~Ye, J.~Huang, X.~Chen, J.~Zhu, T.~Yang, J.~Lu, D.~Du, and J.~Zhou, ``{WebFace260M: A Benchmark Unveiling the Power of Million-Scale Deep Face Recognition},'' in \emph{Proc. IEEE/CVF Conference on Computer Vision and Pattern Recognition}, 2021.

\bibitem{karkkainen2021fairface}
K.~Karkkainen and J.~Joo, ``{FairFace: Face Attribute Dataset for Balanced Race, Gender, and Age for Bias Measurement and Mitigation},'' in \emph{Proc. IEEE/CVF Winter Conference on Applications of Computer Vision}, 2021.

\bibitem{deng2009imagenet}
J.~Deng, W.~Dong, R.~Socher, L.-J. Li, K.~Li, and L.~Fei-Fei, ``Imagenet: A large-scale hierarchical image database,'' in \emph{Proc. IEEE/CVF Conference on Computer Vision and Pattern Recognition}, 2009.

\bibitem{lucy2021gender}
\BIBentryALTinterwordspacing
L.~Lucy and D.~Bamman, ``{Gender and Representation Bias in GPT-3 Generated Stories},'' in \emph{Proc. of the Third Workshop on Narrative Understanding}, 2021. [Online]. Available: \url{https://aclanthology.org/2021.nuse-1.5}
\BIBentrySTDinterwordspacing

\bibitem{he2016deep}
K.~He, X.~Zhang, S.~Ren, and J.~Sun, ``Deep residual learning for image recognition,'' in \emph{Proc. EEE/CVF Conference on Computer Vision and Pattern Recognition}, 2016.

\end{thebibliography}
}
\end{document}